\documentclass{article}
\usepackage{fvextra}  
\usepackage{microtype}
\usepackage{graphicx}
\usepackage{subfigure}
\usepackage{booktabs} 

\usepackage{fvextra}
\usepackage{filecontents}

\usepackage{hyperref}

\usepackage[accepted]{icml2025}

\usepackage{amsmath}
\usepackage{amssymb}
\usepackage{mathtools}
\usepackage{amsthm}

\usepackage[capitalize,noabbrev]{cleveref}

\theoremstyle{plain}
\newtheorem{theorem}{Theorem}[section]

\newtheorem{lemma}[theorem]{Lemma}

\theoremstyle{definition}

\theoremstyle{remark}

\usepackage[textsize=tiny]{todonotes}

\icmltitlerunning{A Voter-Based Stochastic Rejection-Method Framework for Asymptotically Safe Language Model Outputs}

\begin{document}

\twocolumn[
\icmltitle{A Voter-Based Stochastic Rejection-Method Framework for Asymptotically Safe Language Model Outputs}

\begin{icmlauthorlist}
\icmlauthor{Jake Watts}{yyy}
\icmlauthor{Joel Sokol}{yyy}
\end{icmlauthorlist}

\icmlaffiliation{yyy}{Department of Industrial and Systems Engineering, Georgia Institute of Technology, Atlanta, Georgia, USA}
\icmlcorrespondingauthor{Jake Watts}{jwatts60@gatech.edu}

\vskip 0.3in
] 
\makeatletter
\renewcommand{\Notice@String}{}%
\makeatother
\printAffiliationsAndNotice{}
\begin{abstract}
We propose an approach for preventing unsafe or otherwise low-quality large language model (LLM) outputs by leveraging the stochasticity of LLMs, an approach we call Repeated Checking with Regeneration (RCR). In this system, LLM checkers vote on the acceptability of a generated output, regenerating it if a threshold of disapproval is reached, until sufficient checkers approve. Based on our estimators for cost and failure rate and experimental data tailored to the application, our algorithm achieves a desired expected failure rate at Pareto-optimal cost. The failure rate provably decreases exponentially as a function of cost, and the models reasonably estimate the actual performance of such a system in action, even with limited data. This approach does not depend on the language model used, and could allow cheap, small LLMs to control, constrain, or at some tasks even outperform very complex and costly ones.

\end{abstract}

\section{Introduction}
\label{Introduction}
Despite the power and utility of large language models (LLMs), hallucinations and other mistakes remain a major concern. LLMs can effectively deceive \cite{Hagendorff}, give incorrect or incomplete medical advice \cite{Drake}, hallucinate false information \cite{Ji}, be jailbroken and deceived \cite{Yong}, and fail to follow important instructions, especially when users aim to trick them (see Section 3; see also \cite{Richard}). Even at this early stage, the consequences of these errors can be substantial; legal liability may exist when models defame people, create malware, encourage self-harm, cause wrongful death, or assist in criminal activity \cite{Lemley}. In February of 2024, a Civil Resolution Tribunal in British Columbia ruled that Air Canada was liable for a discount its chatbot had mistakenly promised a customer in a misinterpretation of Air Canada’s policy for the discount \cite{Yagoda}. As language models take on more important roles going forward, preventing such mistakes will become increasingly important.

One solution is simply to improve the models, by providing them more training data and more parameters. This is certainly leading to improvements, but it comes with its own drawbacks. Leading models like GPT-5 (at time of writing) are already far too large to be run on household computers, and have been trained on trillions of words. In 2023, it is estimated that OpenAI used over half a Gigawatt-hour per day just on the GPU servers running ChatGPT \cite{Vries}, and companies like Microsoft are looking into producing their own electricity just to be able to power new AI \cite{Alymer}. We propose that our method of using LLMs to preemptively catch their own mistakes provides a cost-efficient and effective alternative or supplementary approach to improving output and preventing dangerous errors, misuse, or misalignment.

Several approaches already exist for using LLMs to catch their own errors. Tree of thought prompting \cite{Long} uses a specific structure building upon chain of thought prompting methods which aim to invoke reasoning in LLMs. Other methods are more obviously analogous to social systems. LLMs are sometimes used to simulate social behavior \cite{Leng} and AutoGen \cite{Wu} and ChatEval \cite{Chan} use multi-agent debate/conversation to improve performance while AutoDefense \cite{Zeng} prevents jailbreaks in a similar manner. \cite{Pang} uses social simulation as a tool for alignment of LLMs with human values.

This paper focuses on a specific paradigm to avoid unsafe or low-quality outputs, Repeated Checking with Regeneration (RCR). This method minimizes the possibility of “group think” or other compounding errors that can occur in human multi-agent systems \cite{Frey}, while still taking advantage of the best outputs from the distribution of random outputs LLMs are capable of generating. We structure the LLM agents in the following way: A generator, which proposes an output, and a number of independent identical (but stochastic, due to a non-zero temperature of the LLM) checkers that assess the quality or safety (or both) of the proposed output and each vote on their approval or disapproval of it. If the number of disapprovals reaches at least some threshold, the generator generates a new, independent output to be checked. This continues until an output is approved. This approach can make bad outputs substantially, even arbitrarily uncommon, greatly improving safety in some applications and quality in others. In particular, we propose a way of choosing the number of checkers $n$ and approval threshold $k$ to achieve Pareto optimality in average cost and failure probability in a specific application and prove a condition under which failure rate can be made arbitrarily low, scaling exponentially with the cost of the method.

As a proof of concept, we demonstrate this approach using an example of a customer service bot given a password and told not to reveal the password to customers. This was inspired by an online game earlier in GPT’s development \cite{Richard} where people competed to trick GPT into revealing a password using the fewest characters, but modified to a more real-world customer service chatbot application.

\section{Model}

For this experiment, we imagine a customer service bot given an administrative passcode that it is told not to reveal to customers. Testing revealed one particular attack prompt that was able to get GPT-3.5-turbo-1106 to reveal the password 22 times out of 50. We test the following approach to improve the security of that chatbot.
After the chatbot generator proposes a response to the attack prompt, $n$ (LLM) checkers each independently judge whether the generated output was acceptable. Each checker is fed the attack and then the generator’s response to the attack before reasoning out loud about its acceptability. The checker’s answer alone is then fed to an (also LLM) evaluator which parses it into a simple “yes” or “no” (meaning approved or disapproved respectively). \emph{This evaluator will likely not be necessary with newer language models, but was the easiest way to get reliably code-readable answers.} If $k$ or more disapprovals are detected, the generator proposes a new response and the process repeats. All prompts used are in the Appendix. For our testing, all LLM instances are of the same or similar level of complexity, GPT-3.5-turbo models, as this task is of an appropriate level of difficulty for that generation of LLM. This approach provides a way to improve output safety/quality without increasing complexity of the chatbot’s neural network, potentially allowing simpler LLMs capable of running on a single PC to outperform more complex LLMs at specific tasks or in terms of safety. In practice, there is no reason the checkers would need to be the same generation or model family as the generator.

In this paper, we demonstrate a successful proof of concept as well as addressing the question of optimal choice of $n$ and $k$ under the tradeoff between cost and failure rate. We demonstrate two empirical estimates used to determine this choice and evaluate the model’s cost-scaling; one systematically underestimates cost and failure rate but requires less data to estimate, and another less biased estimator that requires more testing.

We note that nothing about our method depends in any way on the model family, generation, or problem application. Changing these may lead to different parameters and a different Pareto frontier of checkers and threshold, but the scaling laws and overall method remain the same in any application and with any LLM satisfying the very basic sufficient condition.

It should also be noted that the costs reported here on our test come from using the same generation model for checkers as for the generator. Because the cost at large $n$ is much more sensitive to the cost per check than cost per generator output, these costs would likely be much better if a much cheaper LLM was used for checkers than for generators, especially when chain of thought does not substantially improve checker accuracy.

\subsection{Estimator 1}

The first estimator looks only at the difference in approval between good and bad responses. On our example password problem, it estimates that a failure rate of approximately 0.2\% can be reached at just 7.7 times the cost of generating an output without checking, and one-in-a-trillion at only 41 times the cost of the standard zero-checker system. This estimator, relative to our second estimator, is systematically biased in the direction of overestimating safety, but may be a reasonable heuristic when selecting $n$ and $k$ and is much cheaper to compute and as such might be reasonable to use when stakes are low. This overestimation is largely because some bad responses will have higher approval rates than others and these will be more likely at high $n$ and low $k$ to be output by RCR (which this estimator doesn’t account for).

We call the rate of bad responses $b$, and the approval rates of good and bad responses $a_g$ and $a_b$ respectively, and we estimate the ratio of the cost of running a single check to the cost of generating a single response, which we refer to as the cost ratio $c_r$. Then, the probabilities of a given good or bad generated response being successfully output, respectively, are \[p_g(n,k)=1-\sum_{i=k}^n\binom{n}{k}a_g^i(1-a_g)^{n-i}\]  or \[p_b(n,k)=1-\sum_{i=k}^n\binom{n}{k}a_b^i(1-a_b)^{n-i}.\] The failure rate of this method (RCR), the fraction of checker-approved outputs which are bad, is then \[\frac{bp_b(n,k)}{(bp_b (n,k)+(1-b) p_g (n,k))},\] while the average cost per output (as a multiple of the cost of generating a single proposed response) is \[\frac{1+nc_r}{1-(bp_b (n,k)+(1-b) p_g (n,k))}.\] The failure rate of such a system is an improvement on a chatbot with no checkers as long as the checker is at least slightly more likely to approve of good outputs than bad.

We estimated $b,a_g,a_b$, and $c_r$ by generating 50 responses, 11 of which revealed the password in some way, (so $b=0.22\pm0.06$) then each response was checked by 50 checkers. Checkers approved of bad prompts 101 times, giving $a_b=0.184\pm0.017$, and approved of good prompts 1858 times, giving $a_g=0.9528\pm0.0048$. Calculations from token counts obtained in separate trials give $c_r=1.41$, meaning checking a response and evaluating the check costs about 41\% more (based on current pricing for GPT-3.5-turbo-0125) than generating it to begin with in this case.

Given these parameters, we used a simple enumeration algorithm to determine the values of $n$ and $k$ that achieve a given maximum failure rate at the lowest cost. We call these tuples of $n$ and $k$ dominating as there is no other combination that is simultaneously less expensive on average and has a lower failure rate.

\begin{figure}[ht]
\vskip 0.2in
\begin{center}
\centerline{\includegraphics[width=\columnwidth]{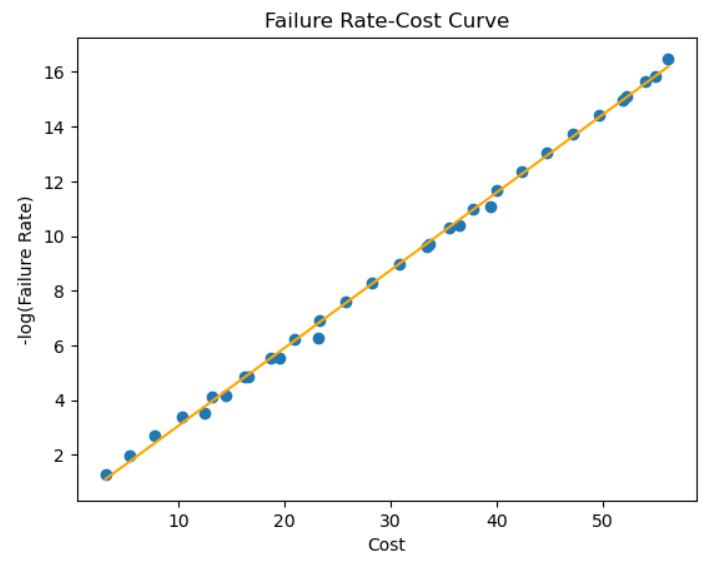}}
\caption{Failure rate-cost curve for Estimator 1. Each point is a Pareto-optimal $(n, k)$; line shows log-linear fit.}
\label{estimator-1}
\end{center}
\vskip -0.2in
\end{figure}

Figure 1 shows the log failure rate/cost curve for this first estimator. We observe a clear linear trend when plotting log (base 10) failure rate, with slope 0.654, meaning that failure rate decreases by a factor of 10 every time cost increases by 3.52 (multiples of the cost of having no checkers). According to this estimator, using the parameters from our 22\% baseline failure rate, the failure rate can be brought as low as one-in-one-trillion at only 41 times the cost of having no checkers. 

\subsection{Estimator 2}

A more conservative and far less biased estimator for determining cost and failure rate takes into account that some bad responses might be much more likely than others to be approved by the checkers and thus be overrepresented in the outputs and produce higher error rates. To take this into account, we use a sample of representative generated responses, some good and some bad, each with its own approval rate and each equally likely to be generated. Checkers then either accept or reject each output based on its approval rate. For this estimator, we used the same 50 generated outputs with 50 independent checks each, and estimated the approval rate of each representative generated response. For each response $j$, let $a_j$ be its approval rate and let $b_j=1$ if the response is bad, and $b_j=0$ otherwise. Let $p_j (n,k)$ be the probability of response $i$ being output (surviving the checkers) when generated.

The overall failure rate of the system is computed by \[\frac{\sum_{j=0}^mb_jp_j(n,k)}{\sum_{j=0}^mp_j(n,k)},\] with cost in the same units as before equal to \[\frac{1+nc_r}{\frac{1}{M}\sum_{j=0}^mp_j(n,k)}.\] Using a similar algorithm as the first estimator, we obtain the dominating tuples which best manage the tradeoff between cost and failure rate.

This estimator should approach the actual cost and failure rate in the limit as the number of sampled generated outputs and used to compute it increases and the accuracy of the estimations of the approval rates of each increases.

\begin{figure}[ht]
\vskip 0.2in
\begin{center}
\centerline{\includegraphics[width=\columnwidth]{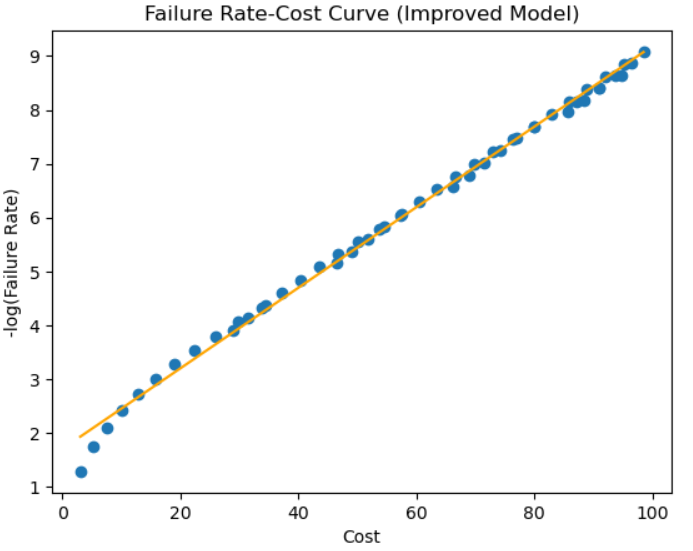}}
\caption{Failure rate-cost curve for Estimator 2. Each point is a Pareto-optimal $(n, k)$; line shows log-linear fit.}
\label{estimator-2}
\end{center}
\vskip -0.2in
\end{figure}

Aside from the first three dominating tuples at low cost, Figure 2 shows a similar log-linear relationship to the first estimator, with failure rate decreasing by a factor of 10 every time cost increases by 13.4. Based on this, a failure rate of one-in-one-trillion is achievable at a cost of only around 160 times the cost of no checkers, compared to the base 0.22 failure rate (for the 50 trials used to estimate this data). 

\section{Results}
To assess the validity of these two failure-rate and cost estimators, we ran an experiment. We set up the proposed system, using GPT-3.5-turbo-1106 to generate prompts (consistent with the data used to derive estimator parameters) and GPT-3.5-turbo-0125 for the checkers and evaluators, with the same prompts each was given when collecting data for the failure rate estimators. We chose $n=6$, $k=4$, as this gave predicted error rates large enough that 1000 trials would be able to detect errors (though this tuple is dominated by better choices). This was run until 1000 responses were approved. In total 1137 responses were generated, with 137 rejected, only one of which was rejected wrongly. Of the 1000 accepted, 32 outputs were unsafe, giving a failure rate of 3.2\% with a 95\% (Wilson) confidence interval of (2.3\%, 4.5\%). This is compared to the 2.2\% predicted by the first estimator, and 4.2\% in the second.

\section{Discussion}
	Both estimators produce a linear relationship between the log of the rate at which bad responses are output and the average cost of producing an output, when dominating $n$ and $k$ are chosen. When tested, the models performed reasonably well at predicting the failure rate and cost of the system despite being based on relatively limited data, particularly the second model. This indicates that, when $n$ and $k$ are chosen effectively according to our estimators (we suggest the second one be used when enough test data can be obtained), this approach may be extremely effective in high-stakes AI safety scenarios.

	This log-linear relationship is powerful, as it allows our approach to achieve very low failure rates without a proportional increase in cost. In cases where the checkers are very poor at discerning between good and bad proposed outputs (the acceptance rates of each are similar on average), the cost increase to reduce failure rate by a factor of 10 can be large. However, when checkers are cheaper than the generator and are reasonably accurate, the cost increase per order of magnitude improvement in failure rate is reasonable, and can even be quite small. As we prove in the appendix, this log-linear relationship for the Pareto frontier is guaranteed in the limit as long as some safe proposed output has a higher approval rate than any unsafe outputs.

	Perhaps most importantly, the behavior of RCR depends much more heavily on the behavior of the checkers than on the behavior of the generators. It does not require that the checker be perfect or even near-perfect either, only that it have an element of randomness so that all checkers do not always agree. This greater relative dependence on checker performance compared to the generator is particularly useful when detecting bad outputs is easier than producing good ones. Since many engineering and design problems fall into this category, this approach may have considerable application, particularly with the possibility of fine-tuning a discriminator to act as the checkers.

\section*{Impact Statement}

This paper presents work whose goal is to advance the capabilities and human alignment of machine learning systems, especially large language models or systems whose outputs can be evaluated by language models. In particular, the ideas of the paper might be effectively used to prevent such systems from taking sets of actions that its creators anticipate the need to prevent. With broad enough instructions, this could prove a significant advancement in alignment or especially control of such systems, or in error prevention of other sorts.

\bibliography{example_paper}
\bibliographystyle{icml2025}

\newpage
\appendix
\onecolumn
\section{Appendix: Proof}
Our proof will follow a sequence of $n_m$, the number of checkers, and $k_m$ the approval threshold, and show that, for some sufficiently large starting point to the sequence $n_0, k_0$ and some appropriate step size in $n$ and $k$, we can guarantee that eventually, the failure probability will decrease at least exponentially to leading order as a function of cost, provided our sufficient condition holds. We do so by proving that, in this sequence's limit, all possible outputs except for the one with the highest approval probability survive with exponentially decreasing probability relative to the most-approved-of output, and that the ratio of these probabilities becomes exponential in $m$ while our cost function grows linearly with $m$.

To prove the sufficient condition for the exponential relationship, suppose that we have some large number of possible outputs, each with different approval rates $p_i$ for each output $i$ in the index set $I$, and probability of being generated $g_i\in (0,1)$. Let the approval rate of the output with the highest approval rate be $p_1$ with generation probability $\gamma_1$, and the approval rate of the next most-approved of output be $p_2$ with generation probability $\gamma_2$ and so on. Assume as our sufficient condition that the strictly most approved-of output is a safe output.

Imagine that we start at some large $n_0$ $k_0$, where $n$ is the number of checkers and $k$ (in this proof) is the number that must approve for the generated result to be output. Define some $\Delta n, \Delta k$ and $\beta:=\frac{\Delta k}{\Delta n}$ such that $\beta \in (\beta', p_1)$ where $\beta':=p_2+\frac{D(p_2||p_1)}{\ln(\frac{p_1(1-p_2)}{p_2(1-p_1)})}$, and where $D(a||b)$ is the Kullback-Lleibler divergence between a and b. Let $n_m=n_0+m\Delta n, k_m =k_0 +m\Delta k$, so that we are moving on a line in $n$\textendash$k$-space. Suppose further that $k_0$ is chosen so that $k_0=\lfloor\beta n_0\rfloor$. Observe that $k_m/n_m=\beta+O(1/m)$ for all positive $m$. We will later use the approximation $k_m\approx\beta n_m$

\begin{theorem}
    For some $n_0, k_0$, let the sequence of the probability of the strictly most-likely-to-be-accepted output being output be $P_{output,m}:=\frac{\gamma_1 Pr[Binom(n_k, p_1)\ge k_m]}{\sum_{i\in I}\gamma_iPr[Binom(n_k, p_i) \ge k_m]}$, and let the cost of generating an output be $C_m:=\frac{n_m+c_r}{\sum_{i\in I}Pr[Binom(n_m, a_i) \ge k_m]}$ Then $\exists m'\in \mathbb{Z}_+, \delta >0$ such that $\ln(\frac{1}{P_{output,m}})\ge \delta C_m \forall m>m'$.
\end{theorem}

To prove this, let us define $P_m^1:=Pr[(Binom(n_m, p_1)> \beta n_m]$ and $P_m^2:=Pr[Binom(n_m, p_2) > \beta n_m]$, and define $S_m:=\frac{n_m}{P_m^1}$. Define the ratio of the output probabilities $R_m:=\frac{P_m^1}{P_m^2}$.

\begin{lemma}
    $\exists m'\in \mathbb{Z}_+, \epsilon>0$ such that $\ln(R_m)\ge\epsilon \frac{n_m}{P_m^1} \;\forall m\in\mathbb{Z}_+, m\ge m'$
\end{lemma}

To see this, note that a Chernoff lower-tail bound tells us that $P_m^1\ge 1-e^{-n_mD(\beta||p_1)}$ where $D(a || p)$ is the Kullback-Leibler divergence between $Bernoulli(a)$ and $Bernoulli(p)$, equal to $a\ln(\frac{a}{p})+(1-a)\ln(\frac{1-a}{1-p})$. As this $D(\beta || p_1)$ does not depend on $n_m$, we can say that, for some $m$ onward, \[P_m^1 =Pr[Binom(n_m, p_1)>k_m]\ge\frac{1}{2}\]. This one half is arbitrary, and we could just as easily have chosen any number in $(0,1)$, and it would still be true because the exponential is decreasing in $n_m$ and therefore in $m$, going to 0 in the limit.

Now, $S_m$ is just $\frac{n_m}{P_m^1}$, and so, for this same $m$ onward, we can say that $S_m=\frac{n_m}{P_m^1}\ge2n_m$. This tells us that, on this sequence, $S_m$ asymptotically scales no faster than proportional to $n_m$. This implies that, for all sufficiently large $m$, we have $\frac{1}{2}\frac{n_m}{P_m^1}\le n_m$.

Now, the log of our ratio $\ln(R_m)=\ln(P_m^1/P_m^2)$, as our Chernoff bound tells us, is \[n_m[D(\beta||p_2)-D(\beta||p_1)]+o(n_m)\]. 
The term in brackets is $\beta\ln(\frac{p_1}{p_2})+(1-\beta)\ln(\frac{1-p_1}{1-p_2})$ from the expression for Kullback-Leibler divergence. It can be shown that this is equivalent to \[c(\beta):=(\beta-p_2)\ln(\frac{p_1(1-p_2)}{p_2(1-p_1)})-D(p_2||p_1)\]

For any $\beta$ strictly between $\beta':=p_2+\frac{D(p_2||p_1)}{\ln(\frac{p_1(1-p_2)}{p_2(1-p_1)})}$ and $p_1$ this $c(\beta)$ term in brackets is positive, as it is strictly increasing as a function of $\beta$ over $(0,1)$ and it is positive at $\beta=p_1$. Then, for $\beta\in(\beta',p_1)$, we have that, for sufficiently large $m$ so that the $o(n_m)$ term is small, $\ln(R_m)\ge n_m\frac{c(\beta)}{2}$.

Then we can say that in the large $m$ limit, for appropriate $\beta\in(\beta',p_1)$, $\ln(R_m)\ge \frac{c(\beta)}{2}n_m \ge \frac{c(\beta)}{4}S_m$, so if $\epsilon:=\frac{c(\beta)}{4}$, we have proved this lemma (or at least showed that some valid $\beta$ satisfies this lemma).

Using this result, we can say that for some $\beta<p_1$ the ratio of probability of the most-likely-to-be-accepted output being output by RCR to the probability of any other output being output by the RCR system goes to infinity faster than $S_m$ (as this ratio is simply the ratio of their acceptance probabilities $R_m$ times the ratio of their generation probabilities, a constant), in the limit as $m\rightarrow\infty$. In the limit as this ratio gets large, the denominator in our expression for $P_{output, m}$ becomes dominated by the probability of the most likely output, and the leading order correction in this denominator will be the 2nd most-accepted output (by the lemma, as it will, for large $m$, dominate every less accepted output). So, if $R_m$ is the acceptance ratio for the most and 2nd most accepted outputs, and $\delta$ is their generation ratio, then
\[P_{output,m}\rightarrow\frac{\gamma_1 Pr[Binom(n_k, p_1)\ge k_m]}{\gamma_1 Pr[Binom(n_k, p_1)\ge k_m] + \delta\frac{1}{R_m}+o(\frac{1}{R_m})}\]

The Failure Rate we care about is $1-P_{output,m}$ which (for appropriate $\beta)$ is 
\[\frac{\frac{1}{R_m}}{P_{output,m}+\frac{1}{R_m}}+o(\frac{1}{R_m})\]
which asymptotes to $\frac{1}{R_m}$. Therefore, the log failure rate asymptotes to $-\ln(R_m)$, which we know via our lemma to exceed $\epsilon \frac{n_m}{P_{output,m}}$ for some $\epsilon>0$. 

Then, we need only note that in the limit, our cost function $C_m:=\frac{n_m+c_r}{\sum_{i\in I}Pr[Binom(n_m, a_i) \ge k_m]}$ as a numerator $n_m+O(1)$ and denominator which cannot exceed $P_{output,m}$.

Then we can say that, for some $\beta\in(q+\frac{D(p_2||p_1)}{\ln(\frac{p_1(1-p_2)}{p_2(1-p_1)})},p_1)$, where $p_2$ is the acceptance rate of the output with the 2nd highest acceptance rate, the (negative) log failure rate scales at least to within a constant factor of the cost in the limit.

In practice, $\beta$ need only exceed that expression where $q$ is the acceptance rate of the most-accepted unsafe output.

As a final note, this proof indicates, as we discovered when plotting the dominating $n, k$ our algorithm produced, that the Pareto-frontier choices of $n$ and $k$ tend to fall on a line in the limit (subject to integer rounding), and in testing, it seemed that linearity occurred almost immediately. In practice, this might greatly simplify the task of finding Pareto-optimal parameters $n$ and $k$.


\section{Appendix: Prompt}
\subsection{Generator}
\VerbatimInput[
  breaklines=true,breakanywhere=true,
  breaksymbolleft={},breaksymbolright={}
]{appendix-generator.json}

\subsection{Checker} 
\VerbatimInput[
  breaklines=true,breakanywhere=true,
  breaksymbolleft={},breaksymbolright={}
]{appendix-checker.json}

\subsection{Evaluator}
\VerbatimInput[
  breaklines=true,breakanywhere=true,
  breaksymbolleft={},breaksymbolright={}
]{appendix-evaluator.json}

\end{document}